\documentclass[11pt,a4paper]{article}
\usepackage[hyperref]{emnlp2020}
\usepackage{times}
\usepackage{latexsym}
\usepackage{amsmath}
\usepackage{amssymb}
\usepackage{tikz}
\usepackage{ctable}
\usepackage{graphicx}
\usepackage{caption}
\usepackage{balance}
\usepackage{microtype}

\aclfinalcopy 

\title{Dynamic Semantic Matching and Aggregation Network for Few-shot Intent Detection}

\author{Hoang Nguyen$^{1}$, Chenwei Zhang$^{2}$, Congying Xia$^{1}$, Philip S. Yu$^{1}$\\
  $^1$ Department of Computer Science, University of Illinois at Chicago, Chicago, IL, USA \\
  $^2$ Amazon, Seattle, WA, USA \\
  \texttt{\{hnguy7,cxia8,psyu\}@uic.edu, cwzhang@amazon.com}
  }
\date{}

\begin{document}
\maketitle
\begin{abstract}
Few-shot Intent Detection is challenging due to the scarcity of available annotated utterances. Although recent works demonstrate that multi-level matching plays an important role in transferring learned knowledge from seen training classes to novel testing classes, they rely on a static similarity measure and overly fine-grained matching components. These limitations inhibit generalizing capability towards Generalized Few-shot Learning settings where both seen and novel classes are co-existent. In this paper, we propose a novel Semantic Matching and Aggregation Network where semantic components are distilled from utterances via multi-head self-attention with additional dynamic regularization constraints. These semantic components capture high-level information, resulting in more effective matching between instances. Our multi-perspective matching method provides a comprehensive matching measure to enhance representations of both labeled and unlabeled instances. We also propose a more challenging evaluation setting that considers classification on the joint all-class label space. Extensive experimental results demonstrate the effectiveness of our method. Our code and data are publicly available \footnote{\url{https://github.com/nhhoang96/Semantic\_Matching}} .
\end{abstract}

\section{Introduction}
Intent Detection (ID) is a crucial task in natural language understanding, whose objective is to extract underlying intents behind the given utterances. The extracted intents could provide further contexts for further downstream Natural Language Processing tasks such as dialogue state tracking or question answering. Unlike traditional text classification, ID is challenging for two main reasons (1) Utterances are usually short and diversely expressed, (2) Emerging intents occur continuously, especially across different domains \cite{liu2019reconstructing}.

Despite recent advances, state-of-the-art ID methods \citep{haihong2019novel, goo2018slot} require a large amount of annotated data to achieve competitive performance. This requirement inhibits models' capability in generalizing to newly emerging intents with no or limited annotations during inference. Re-training or fine-tuning large models on few samples of emerging classes could easily lead to overfitting problems.    

Motivated by human capability in correctly categorizing new classes with only a few examples \citep{lake2011one,gidaris2018dynamic}, few-shot learning (FSL) paradigms are adopted to tackle the scarcity problems of emerging classes. FSL methods take advantage of a small set of labeled examples (support set) to learn how to discriminate unlabeled samples (query samples) between classes, even those not seen during training.

Recent works in FSL \citep{sun-etal-2019-hierarchical, ye-ling-2019-multi} focus on learning the matching information between the labeled samples (support) and the unlabeled samples (query) to provide additional contextual information for instance-level representations, leading to effective prototype representation. However, these methods only extract similarity based on fine-grained word semantics, failing to capture the diverse expressions of users' utterances. This problem could further lead to overfitting either to seen intents or novel intents, especially in the challenging Generalized Few-shot Intent Detection (GFSID) setting \cite{xia2020cg} where both seen and novel intents are existent in a joint label space during inference. Instead, matching support and query samples on coarser-grained semantic components could provide additional informative contexts beyond word levels. For instance, two utterances \textit{"i need to get a table at a pub with southeastern cuisine"} and \textit{``book a spot for six friends"} share a similar intent label \textit{``Book Restaurant"}. While word-level semantics might find similar action words as \textit{``get"} and \textit{``book"}, these words do not necessarily contribute to the correct intent findings. Instead, coarser-grained semantics such as \textit{``get a table"} and \textit{``book a spot"} could provide further hints to identify \textit{``Book Restaurant"} intent.    

As semantic components (SC) could be effectively extracted from multi-head self-attention, matching these SC between support and query can enhance both query and support representations, leading to improvements in generalization from seen training classes to unseen testing classes. To further enhance the dynamics of extracted SC across various domains and diversely expressed utterances, we introduce additional head regularizations. In addition, to overcome the insufficiency of a single similarity measure for matching sentences with diverse semantics, a more comprehensive matching method is further explored.    

Our main contribution is summarized as follows:
\begin{itemize}
    \item We propose a Semantic Matching and Aggregation Network that automatically extracts multiple semantic components from support and query sentences via multi-head self-attention. Additional regularizations are introduced to (1) encourage extracted heads to attend to all words of utterances and (2) encourage semantic alignment between utterances with similar intent labels.
    \item Comprehensive multi-perspective matching is proposed to reduce reliance on a single fixed similarity measure and enhance generalizability towards Generalized Few-shot Learning setting (GFSL).  
    \item We also propose a more challenging but realistic FSL and GFSL evaluation setting. 
\end{itemize}

\section{Related Work}
\label{sec:relatedwork}
\paragraph{Few-shot Learning}
Few-shot learning refers to problems where classifiers are required to generalize to unseen classes with only a few training examples per class \cite{chen2019closerfewshot}. To overcome challenges of potential overfitting, most FSL methods adopt meta-learning approach where knowledge is extracted and transferred across multiple tasks. There are two major approaches towards FSL: (1) metric-based approach whose goal is to learn feature extractor that extract and generalize to emerging classes \citep{vinyals2016matching, snell2017prototypical, sung2018learning}, and (2) optimization-based approach that aims to optimize model parameters from few samples \citep{santoro2016meta, finn2017model, DBLP:conf/iclr/RaviL17, DBLP:conf/iclr/MishraR0A18}. In this work, we focus mostly on metric-based learning approach. Specifically, we extend Prototypical Network (PN) \cite{snell2017prototypical} in which prototypes are not only represented by support samples but also matching information between support and query samples.

Traditionally, FSL methods are evaluated in episodic procedure due to the major principle that test and train conditions must match \cite{vinyals2016matching}. Each episode represents a meta-learning task in which the models explicitly ``learn to learn" minimize the loss on an unlabeled/ query set given the support/ labeled set. However, we claim that this evaluation is lack of practicality for two main reasons. First, evaluation on random samples could not help us understand the strengths or weaknesses of the model. For instance, if the trained model overfits a subset of novel classes, it is  impossible to pinpoint the overfitting classes with episodic evaluation. Secondly, in realistic applications, there is a need to categorize unlabeled samples into one of the novel/joint classes, rather than a set of sampled classes. Episodic  testing  does not provide an end-to-end systematic evaluation. Therefore, in our work, we propose a more challenging but realistic non-episodic evaluation setting where unlabeled samples are only inferred once with a probablility distribution over a fixed set of classes in novel or joint label space.

\paragraph{Sentence Matching}
Recent FSL works adopt multi-level matching and aggregation methods to improve FSL performance \citep{gao2019hybrid, sun-etal-2019-hierarchical, ye-ling-2019-multi}. Instead of constructing prototypes purely from support samples, recent works integrate matching information between support and query samples on multiple levels. \citet{gao2019hybrid} introduces feature-level and instance-level attention. \citet{sun-etal-2019-hierarchical} introduces additional word-level attention and proposes more advanced multi-cross attention on instance-level. On the other hand, \citet{ye-ling-2019-multi} adopts soft matching between support and query samples to build local context representation for both support and query samples. These methods have been proven effective in few-shot relation classification tasks. However, they rely on overly fine-grained level matching which potentially causes overfitting problems towards either seen or unseen set of classes. Our work mainly differs in two aspects: (1) Comprehensive multi-perspective matching for information matching and (2) Matching on coarser-grained semantic-component levels that are extracted dynamically for effective knowledge transfer, especially in GFSL settings.  
\section{Problem Formulation}
In this section, we provide definitions for both Few-shot Intent Detection (FSID) and GFSID task. Traditional FSL task is defined as C-way K-shot classification task in which classifier performs a series of tasks during both training and inference, which involves C randomly chosen classes with only K labeled samples from each class ($K \leq 5$). These $C \cdot K$ samples are named as support samples. This series of tasks are repeated via episodes \cite{vinyals2016matching}. In each episode, the objective is to correctly classify unlabeled samples (query samples) by using only the support samples.

We denote seen label space as $Y_{s}$, novel label space as $Y_{n}$, and $Y_{s} \cap Y_{n} = \varnothing$. Given the seen labels ($Y_{s})$, we define $D_{s}=\{ (x_1,y_1), (x_2,y_2), ... (x_{N_{s}}, y_{N_{s}}) \}$,  where $N_{s}$ denotes the total number of seen samples and $(x,y)$ denotes a pair of utterance and intent label. Similarly, $D_{n}=\{ (x_1,y_1), (x_2,y_2), ... (x_{N_{n}}, y_{N_{n}}) \}$. 

Given an unlabeled utterance x, the objective of FSID is to maximize correct prediction for x within the novel label subspace $Y_{n}$ as summarized in \eqref{eq:fslobj}.
 \vspace*{-0.3cm}
\begin{equation}
    \hat{y} = \underset{y \in Y_{n}}{arg max} \;p(y|x, D_{n})
    \label{eq:fslobj}
\end{equation}
For GFSID, there exists an additional joint label space $Y_{j} = Y_{s} \cup Y_{n}$. Unlike FSID, GFSID is more challenging as the test samples could come from either seen or novel sample space. The objective function is modified as follows.
 \vspace*{-0.2cm}
\begin{equation}
    \hat{y} = \underset{y \in Y_{j}}{arg max} \;p(y|x, D_{j})
    \label{eq:gfslobj}
\end{equation}
 \vspace*{-0.2cm}
\begin{figure}[t]
\centering
\includegraphics[page=1,scale=0.35]{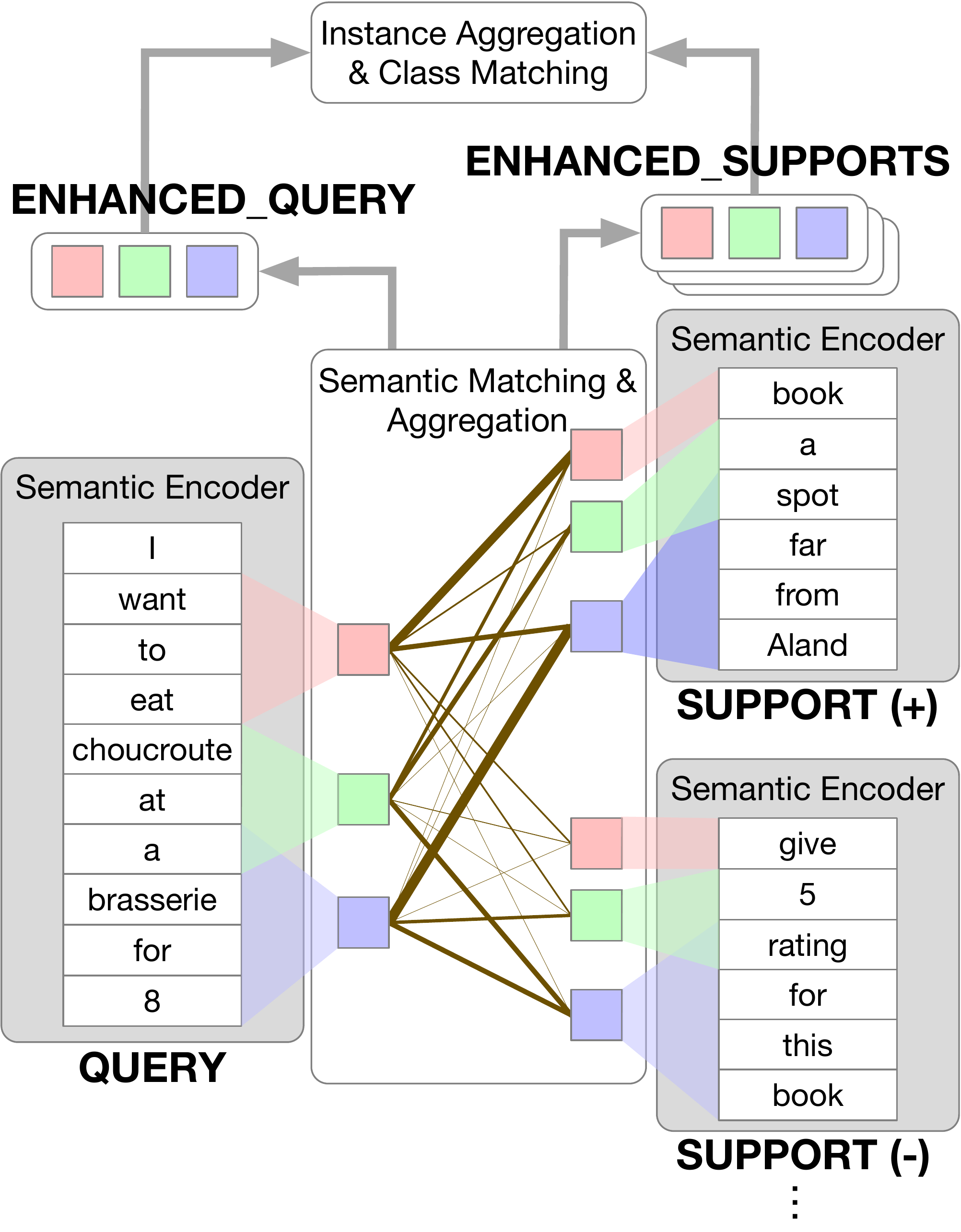}
\caption{Illustration of the proposed Semantic Matching and Aggregation Model for few-shot intent detection. Semantic Components extracted from Semantic Encoder capture high-level semantics beyond word semantic level. Matching these components is more effective than word-by-word matching as contextual phrases are further taken into consideration and non-essential words do not distract the matching functions.}
\label{fig:model_intro}
\end{figure}
\vspace*{-0.4cm}
\section{Methodology}
In this section, we introduce our proposed architecture. Specifically, we divide the framework into 3 main components: Semantic Encoder, Semantic Matching \& Aggregation, Instance Aggregation \& Class Matching as illustrated in Figure \ref{fig:model_intro}.
\vspace*{-0.2cm}
\subsection{Semantic Encoder}
\label{sec:semantic_encoder}
The objective of Semantic Encoder (SE) is to extract semantic components from the given support or query instances. Given an input support or query instance $\mathbf{x}=[x_1, x_2,..., x_T]$ with T words, SE first maps each word into a $d_w$ dimensional word embedding. Pre-trained embedding such as Glove \cite{pennington2014glove}, or even contextualized embedding BERT \cite{devlin2018bert} could be elevated. In our work, we adopt pre-trained FastText embedding \cite{bojanowski2017enriching}.

To capture semantic and syntactic information of the given instance, we adopt self-attentive semantic encoder inspired by multi-head self-attention in \cite{lin2017structured}. Specifically, we first use Bi-Directional Long short-term Memory (Bi-LSTM) to capture contextual information between words within a sentence.
\begin{equation}
\begin{aligned}
    & \overrightarrow{\mathbf{h}_t} = \overrightarrow{{LSTM}}(\mathbf{w}_t, \overrightarrow{\mathbf{h}_{t-1})} \\
    & \overleftarrow{\mathbf{h}_t} = \overleftarrow{{LSTM}}(\mathbf{w}_t, \overleftarrow{\mathbf{h}_{t+1})}
    \end{aligned}
\end{equation}
The hidden representation of $\mathbf{x}$ (denoted as $\mathbf{H} \in \mathbb{R}^{T \times 2d_h}$) is a concatenation of both forward and backward hidden states where $d_{h}$ is the hidden size. 
\begin{equation}
    \mathbf{H} = [\mathbf{h}_{1}, \mathbf{h}_{2}, ..., \mathbf{h}_{T}]
    \label{eq:hidden_state}
\end{equation}
To capture more fine-grained signals other than sentence vector representation, self-attention mechanism is adopted to extract important semantic components of the sentence. Each semantic component, denoted as ``head", is learned from the hidden state $\mathbf{H}$ via multi-layer perceptrons (MLP).
\begin{equation}
   \mathbf{A} = softmax(\mathbf{W}_{s2}tanh(\mathbf{W}_{s1}\mathbf{H}^T))
\end{equation}
where $\mathbf{W}_{s1}$, $\mathbf{W}_{s2}$ are the learning weights with dimension of $ \mathbb{R}^{\;{d_a \times 2d_{h}}}$ and $\mathbb{R}^{\; r \times d_{a}}$. $d_a$ and $r$ can be simply seen as the hidden size and output size of the embedded feed-forward network. $r$ represents the number of heads or important features that the network extracts from the given sentence. The r-head representation $\mathbf{M} \in \mathbb{R}^{\;{r \times 2{d_h}}}$ is a product of attention matrix and the obtained hidden states $\mathbf{M}= \mathbf{A} \mathbf{H}$.

Additional regularization terms are introduced to enforce (1) Each head focuses on different aspects of a sentence, (2) All words in an utterance are covered by the extracted heads, (3) Head distribution between query and support with the same intent labels should be similar to one another. These regularized terms are optimized together with the query classification loss ($\mathcal{L}_{class}$) to further improve the model's performance. In summary, our training loss is summarized as follows.
\begin{equation}
    \mathcal{L} = \mathcal{L}_{class} + \alpha \mathcal{L}_{self\_attn} + \beta \mathcal{L}_{uniform} + \gamma \mathcal{L}_{discr}
\end{equation}
where $\alpha$, $\beta$, $\gamma$ are hyperparameters.

\paragraph{Self-attention regularization}
 Additional regularization term is needed to enforce that each attention head focuses on different semantic components of the utterance. The most intuitive approach is to minimize the number of ``attended" tokens for each head, forcing each head vector to attend to a single aspect of the given sentence \cite{lin2017structured}.
\begin{equation}
    \mathcal{L}_{self\_attn} = ||(\mathbf{A} \mathbf{A}^T -\mathbf{I})||_F^2
\end{equation}
where $\mathbf{A}$ denotes the obtained attention matrix from SE and $||\bullet||_F^2$ denotes Frobenius matrix norm.
\paragraph{Head uniform regularization}
To ensure that all words of a given utterance are covered by at least one head obtained by multi-head self-attention, we minimize the Kullback-Leiber (KL) divergence between the word probability distribution over all heads ($\sum_{i=1}^{r}\mathbf{A}_i$) and a uniform distribution U. 
 \vspace*{-0.3cm}
\begin{equation}
    \mathcal{L}_{uniform} = D_{KL} (p(\sum_{i=1}^{r}\mathbf{A}_i)||U)
\end{equation}

Head uniform regularization is introduced to increase robustness and dynamic of extraction behavior by covering even rare words that are not widely used in utterances. 

\paragraph{Head distribution regularization}
To encourage semantic alignment between support and query samples of the same intent, we minimize the KL divergence in terms of head distributions among those with similar intents while maximizing KL divergence among those that are different.
\vspace*{-0.3cm}
\begin{equation}
    \begin{aligned}
    \mathcal{L}_{discr} = (\hat{Y}_{Q} = Y_{S}) D_{KL} (p(\sum_{i=1}^{L_Q}\mathbf{A}_Q) || p(\sum_{j=1}^{L_S}\mathbf{A}_S))  \\
    - (\hat{Y}_{Q} \neq Y_{S}) D_{KL} (p(\sum_{i=1}^{L_Q}\mathbf{A}_Q) || p(\sum_{j=1}^{L_S}\mathbf{A}_S))
        \end{aligned}
\end{equation}
$L_Q$ and $L_S$ denote the lengths of query and support sentences respectively. $\hat{Y}_Q$ and $Y_S$ denote predicted query label and ground truth support label respectively. This regularization allows for dynamic multi-head self-attention extraction behavior by incorporating query predicted label from downstream task into the objective function. 

\subsection{Semantic Matching \& Aggregation}
In order to enrich representations for both support and query instances, given SCs extracted from Semantic Encoder, we introduce Semantic Matching \& Aggregation module to capture and aggregate matching local contexts between support and query via SCs. Specifically, our module is made up of two components: (1) Multi-perspective Semantic Matching and (2) Semantic Aggregation. 

Extracted head representations from SE (matrix $\mathbf{M}$) for both support and query samples are used in this module.We denote representations of k-th support sample as $\mathbf{S}_k=[\mathbf{M}_{s_k}^1, \mathbf{M}_{s_k}^2,...,\mathbf{M}_{s_k}^r]$ and query sample as $\mathbf{Q}= [\mathbf{M}_q^1, \mathbf{M}_q^2,...,\mathbf{M}_q^r]$  respectively , where r denotes the number of extracted heads from SE. This module is applied to both support and query samples to build an enhanced instance representation $\hat{\mathbf{S}}_{k}$ and $\hat{\mathbf{Q}}$. For simplicity, we only define the one-way matching ($\mathbf{S}_{k} \rightarrow \mathbf{Q}$).
\subsubsection{Multi-perspective Semantic Matching}
Following \cite{ijcai2017-579}, we define the multi-perspective matching function $f_m$ between two vectors as $\mathbf{m} = f_m(\mathbf{v}_1, \mathbf{v}_2;\mathbf{W})$ where $\mathbf{W} \in \mathbb{R}^{\;l \times d}$ is a trainable weight parameter. \textit{l} is a hyperparameter defining the number of perspectives. Each perspective in vector $\mathbf{m}$ is a cosine similarity between weighted vectors $\mathbf{v}_1$ and $\mathbf{v}_2$. In other words, $m_k = cosine(\mathbf{W}_k \circ \mathbf{v}_1, \mathbf{W}_k \circ \mathbf{v}_2)$ where $\circ$ defines element-wise multiplication. 

We define four different components of multi-perspective matching method as follows.
\paragraph{Head-wise Matching} Each head's forward and backward contextualized embedding  of $\mathbf{S}_{k}$ are compared with the corresponding head's forward and backward contextual embedding of $\mathbf{Q}$. 
\vspace*{-0.2cm}
\begin{equation}
\begin{aligned}
& \overrightarrow{\mathbf{m}_i^{head\_wise}} = f_{m}(\overrightarrow{\mathbf{M}_{s_k}^i}, \overrightarrow{\mathbf{M}_{q}^i};\mathbf{W}^1) \\
& \overleftarrow{\mathbf{m}_i^{head\_wise}} = f_{m}(\overleftarrow{\mathbf{M}_{s_k}^i}, \overleftarrow{\mathbf{M}_{q}^i};\mathbf{W}^2)
\end{aligned}
\end{equation}
\paragraph{Max-pooling Matching} Each head's forward and backward contextualized embedding  of $\mathbf{S}_{k}$ is compared with all heads' forward and backward contextual embedding of $\mathbf{Q}$. However, only the maximum value in each dimension is extracted and retained in the matching vector.
\vspace*{-0.2cm}
\begin{equation}
\begin{aligned}
& \overrightarrow{\mathbf{m}_i^{max}} = \underset{j \in (1..r)}{max} f_{m}(\overrightarrow{\mathbf{M}_{s_k}^i}, \overrightarrow{\mathbf{M}_{q}^j};\mathbf{W}^3) \\
& \overleftarrow{\mathbf{m}_i^{max}} = \underset{j \in (1..r)}{max} f_{m}(\overleftarrow{\mathbf{M}_{s_k}^i}, \overleftarrow{\mathbf{M}_{q}^j};\mathbf{W}^4)
\end{aligned}
\end{equation}
\paragraph{Attentive Matching}
Unlike Max-Pooling matching, Attentive Matching is divided into two steps (1) Head representative is aggregated via similarity scores between different heads of each support and query sample (2) Matching head representative and the support heads. For similarity measure, cosine function is utilized.
\vspace*{-0.2cm}
\begin{equation}
\begin{aligned}
& \overrightarrow{\mathbf{\beta}_{i,j}} = cosine(\overrightarrow{\mathbf{M}_{s_k}^i}, \overrightarrow{\mathbf{M}_{q}^j}) \\
& \overleftarrow{\mathbf{\beta}_{i,j}} = cosine(\overleftarrow{\mathbf{M}_{s_k}^i}, \overleftarrow{\mathbf{M}_{q}^j})
\end{aligned}
\end{equation}
Head representative is defined as a weighted sum of all query heads.
\vspace*{-0.2cm}
\begin{equation}
\begin{aligned}
& \overrightarrow{\mathbf{M}_i^{rep}} = \frac{\sum_{j=1}^r \overrightarrow{\mathbf{\beta}_{i,j}} \cdot \overrightarrow{\mathbf{M}_{q}^j} }{\sum_{j=1}^r \overrightarrow{\mathbf{\beta}_{i,j}}} \\
& \overleftarrow{\mathbf{M}_i^{rep}} = \frac{\sum_{j=1}^r \overleftarrow{\mathbf{\beta}_{i,j}} \cdot \overleftarrow{\mathbf{M}_{q}^j} }{\sum_{j=1}^r \overleftarrow{\mathbf{\beta}_{i,j}}}
\end{aligned}
 \label{eq:att_rep}
\end{equation}
The computed head representative is compared with each head's contextualized embedding of $\mathbf{S}_k$.
\vspace*{-0.2cm}
\begin{equation}
\begin{aligned}
    & \overrightarrow{\mathbf{m}_i^{attn}} = f_{m}(\overrightarrow{\mathbf{M}_{s_k}^i}, \overrightarrow{\mathbf{M}_{i}^{rep}};\mathbf{W}^5) \\
    & \overleftarrow{\mathbf{m}_i^{attn}} = f_{m}(\overleftarrow{\mathbf{M}_{s_k}^i}, \overleftarrow{\mathbf{M}_{i}^{rep}}; \mathbf{W}^6)    
    \end{aligned}
\end{equation}
\paragraph{Max-Attentive Matching}
Similar to Attentive Matching, Max-Attentive extracts head representative in Equation \eqref{eq:att_rep}. Instead of doing the pairwise matching, Max-Attentive conducts max-pooling between $\mathbf{M}_{j}^{rep}$ and $\mathbf{M}_{s_k}^i$.
\vspace*{-0.2cm}
\begin{equation}
\begin{aligned}
& \overrightarrow{\mathbf{m}_i^{max\_attn}} = \underset{j \in (1..r)}{max} f_{m}(\overrightarrow{\mathbf{M}_{s_k}^i}, \overrightarrow{\mathbf{M}_{j}^{rep}};\mathbf{W}^7) \\
& \overleftarrow{\mathbf{m}_i^{max\_attn}} = \underset{j \in (1..r)}{max} f_{m}(\overleftarrow{\mathbf{M}_{s_k}^i}, \overleftarrow{\mathbf{M}_{j}^{rep}};\mathbf{W}^8)
\end{aligned}
\end{equation}

\subsubsection{Semantic Aggregation}
In order to aggregate the matched representation into a single instance representation, we use another Bi-LSTM whose input is a concatenation of matched representation in previous sections.
\begin{equation}
    \resizebox{\columnwidth}{!}{ 
    $
        \begin{aligned}
            & \overrightarrow{\hat{\mathbf{S}}_{k}} = LSTM(\overrightarrow{\mathbf{m}_i^{head\_wise}} \oplus \overrightarrow{\mathbf{m}_i^{max\_attn}} \oplus \overrightarrow{\mathbf{m}_i^{attn}} \oplus \overrightarrow{\mathbf{m}_i^{max}}) \\
            & \overleftarrow{\hat{\mathbf{S}}_{k}} = LSTM(\overleftarrow{\mathbf{m}_i^{head\_wise}} \oplus \overleftarrow{\mathbf{m}_i^{max\_attn}} \oplus \overleftarrow{\mathbf{m}_i^{attn}} \oplus \overleftarrow{\mathbf{m}_i^{max}}) 
        \end{aligned}
        $
}
\end{equation}
where $\oplus$ denotes concatenation operation.

Similarly, we obtain the final representation of query with reverse matching ($\mathbf{Q} \rightarrow \mathbf{S}_{k}$) where $\{\mathbf{\hat{Q}}, \mathbf{\hat{S}}_k\} \in \mathbb{R}^{2d_h}$.
\subsection{Instance Aggregation \& Class Matching }
As indicated in previous works, when class label covers diverse semantics, each support instance contributes differently to the class prototype given the query instance. Therefore, we replace the mean operation over all support instances of PN with attentive aggregation. Attention weight for each support instance $\mathbf{\hat{S}}_k$ is learned via a MLP.
\vspace*{-0.2cm}
\begin{equation}
    \alpha_k = \mathbf{W}_{9}^{T}(ReLU(\mathbf{W}_{10}[\mathbf{\hat{S}}_k \oplus \hat{\mathbf{Q}}]))
    \label{eq:instanceaggr}
\end{equation}
Support prototype ($\mathbf{\hat{S}}$) is computed as a weighted sum aggregation via support attention weight and each k-th support instance representation.
\vspace*{-0.2cm}
\begin{equation}
    \mathbf{\hat{S}} = \sum_{k=1}^K softmax(\alpha_k) \mathbf{\hat{S}}_k
\end{equation}

Another MLP is used as class matching function by using support prototype and query representation.
\vspace*{-0.2cm}
\begin{equation}
    \hat{Y} = \mathbf{W}_{9}^{T}(ReLU(\mathbf{W}_{10}[\mathbf{\hat{S}} \oplus \mathbf{\hat{Q}}]))
    \label{eq:classmatch}
\end{equation}
Weights $\mathbf{W}_9 \in \mathbb{R}^{d_h}$ and $\mathbf{W}_{10}\in \mathbb{R}^{d_h \times 4d_h}$ are shared between instance aggregation (Equation \eqref{eq:instanceaggr}) and class matching (Equation \eqref{eq:classmatch}) for optimal performance \cite{ye-ling-2019-multi}.
\section{Experiments}
\begin{table}[t]
\centering
\caption{Details of SNIPS and NLUE (Fold 1) datasets.}
\resizebox{\columnwidth}{!}{%
\begin{tabular}{|c|c|c|}
\hline 
& SNIPS & NLUE \\
\hline 
\# Seen classes ($|Y_s|$) & 5 & 48  \\
\# Novel classes ($|Y_n|$)& 2 & 16 \\
\# Seen samples ($N_s$) & 7887 & 6393 \\
\# Novel samples ($N_n$) & 769 & 274 \\
\# Joint samples ($N_j$) & 2688 & 1873 \\
\# Seen samples per class ($\bar{N}_s$)  & 1577.4 & 133.2 \\
\# Novel samples per class ($\bar{N}_n$)   & 384.5 & 17.1 \\
\# Joint samples per class ($\bar{N}_j$)   & 384.0 & 29.3 \\
\hline
\end{tabular}%
}
\label{detailsdataset}
\end{table}
\vspace*{-0.1cm}
\subsection{Dataset}
We evaluate our proposed model on two real-world datasets for the GFSID task: SNIPS-NLU (SNIPS) and NLU-Evaluation Dataset (NLUE). Both datasets are widely as benchmarks for Natural Language Understanding tasks. Statistics of both datasets are summarized in Table \ref{detailsdataset}. 

For each dataset, we define Seen-Novel-Joint datasets. To build a joint dataset ($D_{j}$), we aggregate 20\% of seen intent utterances with novel intent utterances. The remaining seen intent utterances (80\%) are used as training data (reported $N_{s}$ in Table \ref{detailsdataset}). The support samples (1 or 5 shots) are randomly sampled in advance and not counted in either $N_{s}$, $N_{n}$ or $N_{j}$.

\textbf{SNIPS-NLU}: Following \cite{xia2018zero},we select two intents (RateBook and AddToPlaylist) as novel/ emerging intents and the other five intents as seen intents. 

\textbf{NLUE}: Following \cite{liu2019benchmarking}, we utilize a subset of utterances covering 64 intents. We randomly choose 16 intents as unseen intents while the remaining 48 intents are considered seen.
\subsection{Baselines}
We compare our model with several traditional FSL models, and specifically metric-based network models. For fair comparison and consistency, we implement our SE proposed in Section \ref{sec:semantic_encoder} for all considered baselines. Final instance embedding is obtained as a mean operation over all heads. The only exception is HAPN and MLMAN as they require local matching (i.e. word matching) modules. In that case, we use output of Bi-LSTM (in Equation \eqref{eq:hidden_state}) and enhance it with the head regularization term (Section 4.1) during training.
\begin{itemize}
    \item \textbf{Matching Network (MN)} \cite{vinyals2016matching}: few-shot learning paradigm mapping samples to labels via attention mechanism.
    \item \textbf{Prototypical Network (PN)} \cite{snell2017prototypical}: few-shot method categorizing samples via Euclidean distance from class prototypes.
    \item \textbf{Relation Network (RN)} \cite{sung2018learning} few-shot model that uses neural network to learn deep metric known as relation scores.
    \item \textbf{Hybrid Attention-based Prototypical Network (HATT)} \cite{gao2019hybrid}: initial few-shot learning model that integrates feature-level attention and instance-level attention between support and query samples.
    \item \textbf{Hierarchical Prototypical Network (HAPN)} \cite{sun-etal-2019-hierarchical}: few-shot learning  paradigm that extracts similarity on all feature, word and instance levels.

    \item \textbf{Multi-level Matching and Aggregation Network (MLMAN)} \cite{ye-ling-2019-multi}: multi-level matching approach exploiting both fusion and dot product similarity on local/ word level to enhance instance representation.
\end{itemize}
\vspace*{-0.5cm}
\begin{table}[t]
\centering
\caption{Hyperparameters for both datasets.}
\resizebox{\columnwidth}{!}{%
\begin{tabular}{|c||c|c|c|c|c|c|c|c|c|c|c|c||}
\hline 
 & $d_a$ & $d_h$ & r & L & $\alpha$ & $
 \beta$ & $\gamma$ \\
\hline
SNIPS & 20 & 64 & 4 & 5 & 0.0001 & 1e-5 & 0.01 \\
\hline 
NLUE & 20 & 64 & 4 & 5 & 1e-5 & 1e-5 & 0.001 \\
\hline
\end{tabular}%
}
\label{hyperparameter}
\end{table}
\begin{table*}[htb]
\centering
\caption{Experimental result on SNIPS dataset.}
\resizebox{\textwidth}{!}{%
\begin{tabular}{|c||c|c|c||c|c|c||c|c|c||c|c|c||}
\hline 
  & \multicolumn{6}{c||}{\textbf{1-shot}} & \multicolumn{6}{c||}{\textbf{5-shot}} \\
 \hline
 & \multicolumn{3}{c||}{\textbf{Non-episodic (noneps)}} & \multicolumn{3}{c||}{\textbf{Episodic (eps) }} & \multicolumn{3}{c||}{\textbf{Non-episodic (noneps)}} & \multicolumn{3}{c||}{\textbf{Episodic (eps)}} \\
\specialrule{.1em}{0.05em}{.05em}
 Model & S-J & S-N & h\_acc & S-J & S-N & h\_acc & S-J & S-N & h\_acc & S-J & S-N & h\_acc \\
\specialrule{.1em}{0.05em}{.05em}
MN & 73.5 & 86.99 & 79.68 & 82.67 & 85.97 & 84.29 
& 77.31 & 90.12 & 83.22 & 84.6 & 90.12 & 87.27 \\

 PN & 71.61 & 94.67 & 81.54 &87.04 & 89.91 & 88.45 & 85.31 & 93.11 & 89.04 & 91.05 & 92.96 & 92.00 \\

RN & 74.94 & 88.14 & 81.01 & 85.63 & 87.63 & 86.62 & 64.09 
&87.99 & 74.16 & 79.25 & 83.86 & 81.49 \\
 HATT & 71.54 & \textbf{93.76} & 81.16 & 84.51 & \textbf{93.55} & 88.80 &
 \textbf{86.53} & 94.15 & \textbf{90.18} & \textbf{91.85} & 93.98 & \textbf{92.90}\\

 MLMAN & \textbf{78.61} & 94.41 & \textbf{85.79} & \textbf{87.77} & 92.48 & \textbf{90.06} &
 79.58 & \textbf{95.06} & 86.64 & 89.27 & 94.13 & 91.64 \\
 HAPN & 74.33 & 91.42 & 81.99 & 85.37 & 91.52 & 88.34 &
86.19 &  92.85 & 89.40 &  89.4 & \textbf{94.32} & 91.79 \\
\hline

 \textbf{Ours} & \textbf{81.85} & \textbf{95.84} & \textbf{88.29} & \textbf{88.1} & \textbf{95.48} & \textbf{91.64} 
 & \textbf{87.87} & \textbf{97.01} & \textbf{92.21} & \textbf{93.18} & \textbf{96.81} & \textbf{94.96} \\ 
\hline
\end{tabular}%
}
\label{snipsresult}
\end{table*}
\begin{table*}[htb]
\centering
\caption{Experimental result on NLUE dataset.}
\resizebox{\textwidth}{!}{%
\begin{tabular}{|c||c|c|c||c|c|c||c|c|c||c|c|c||}
\hline 
  & \multicolumn{6}{c||}{\textbf{1-shot}} & \multicolumn{6}{c||}{\textbf{5-shot}} \\
 \hline
 & \multicolumn{3}{c||}{\textbf{Non-episodic (noneps)}} & \multicolumn{3}{c||}{\textbf{Episodic (eps) }} & \multicolumn{3}{c||}{\textbf{Non-episodic (noneps)}} & \multicolumn{3}{c||}{\textbf{Episodic (eps)}} \\
 \hline
 Model & S-J & S-N & h\_acc & S-J & S-N & h\_acc & S-J & S-N & h\_acc & S-J & S-N & h\_acc \\
\hline
 MN & 62.3 & 35.4 & 45.15 & 76.21 & 58.16 & 65.97 & 
 56.27 & 52.55 & 54.35 & 78.85 & 73.69 & 76.18 \\
 
 PN & 62.63 & 36.86 & 46.41 & 80.78 & 58.44 & 67.82 &
 66.2 & 59.49 & 62.67 & 85.13 & 79.39 & 82.16 \\
 
 RN & 56.75 & 27.74 & 37.26 & 73.57 & 49.47 & 59.16 &
 46.5 & 34.31 & 39.49 & 75.23 & 62.15 & 68.07 \\
 
 HATT & \textbf{64.01} & 34.67 & 44.98 & 81.39 & 58.47 & 68.05 &
 67.86 & 61.15 & 64.33 & 78.41 & 74.74 & 76.53 \\
 
 MLMAN & 63.12 &41.61 & \textbf{51.60} & \textbf{82.65} & 60.64 & 69.95 & 
 60.7 & 59.49 & 60.09 & 84.45 & 76.7 & 80.39 \\
 
 HAPN & 60.44 & \textbf{41.78} & 49.41 & 82.00 & \textbf{62.39} & \textbf{70.86} &
 \textbf{68.34} & \textbf{64.6} & \textbf{66.42} & \textbf{84.75} & \textbf{80.11} & \textbf{82.36} \\
 \hline
 \textbf{Ours} & \textbf{66.1} & \textbf{44.11} & \textbf{52.91} & \textbf{89.54} & \textbf{62.81} & \textbf{73.83} 
 & \textbf{72.18} & \textbf{66.96} & \textbf{69.47} & \textbf{87.76} & \textbf{81.12} & \textbf{84.31} \\ 
\hline
\end{tabular}%
}
\label{nlueresult}
\end{table*}
\subsection{Implementation Details}
We use 3-fold cross-validation to tune all of the hyperparameters based on S-J accuracy  on SNIPS and Fold 1 of NLUE datasets as summarized in Table \ref{hyperparameter}. Pre-trained FastText word embedding is used to initialize word embedding and stays fixed during both training and testing for fair comparison between our proposed model and baselines. We train each model over 1000 randomly sampled episodes with learning rate of 0.0001. The number of query samples ($N_Q$) for each episode is 20.

Following \cite{shi2019relational}, we evaluate our models on overall Seen-Joint (S-J) and Seen-Novel (S-N) accuracy. Reported S-J accuracy denotes GFSID evaluation result while S-N indicates traditional FSID results. Reported h-accuracy is a harmonic mean between S-J and S-N accuracy to evaluate the stability of the overall model in both GFSID and FSID settings. 
\paragraph{Episodic Evaluation} Traditional FSL methods are evaluated in episodes due to the major principle that test and train conditions (C-way K-shot) must match \cite{vinyals2016matching}. On SNIPS dataset,we conduct experiments with $K=\{1,5\}$  and $C=2$ with 5 random seed initialization and report average accuracy in Table \ref{snipsresult}. For NLUE dataset, we average accuracy over 10 Folds with similar K and $C=5$. The sampling procedure for GFSL is conducted in a similar way as \citep{shi2019relational}.
\vspace*{-0.1cm}
\paragraph{Non-episodic Evaluation} As mentioned in Section \ref{sec:relatedwork}, Episodic Evaluation is lack of practicality and does not provide an end-to-end system evaluation. Therefore, we also evaluate the models on our proposed non-episodic procedure where unlabeled samples are only inferred once and the predicted probability distribution is over all $Y_{n}$ or $Y_{j}$ label space.
\vspace*{-0.1cm}
\subsection{Experimental Results}
As we observe from Table \ref{snipsresult} and \ref{nlueresult}, our proposed model outperforms the previous baselines by a large margin in both episodic and non-episodic evaluations on both datasets. Our model also observes a consistent stability between FSID and GFSID tasks across both datasets. 

All of the models observe a major decrease in accuracy when evaluated on our challenging non-episodic evaluation as compared to the traditional episodic procedure. Specifically, GFSID tasks are mostly affected by non-episodic evaluation (around 10\% S-J accuracy drop in both datasets). On SNIPS dataset, since both non-episodic and episodic evaluations on S-N are conducted as 2-way 1-shot or 2-way 5-shot, the reported accuracy is almost similar. However, on the other hand, as C and $|Y_{n}|$ or $|Y_{j}|$ are different (5 vs 16 or 64) on NLUE dataset, we observe significant differences in reported S-N accuracy across all models.

On NLUE dataset, S-N accuracy is consistently lower than S-J accuracy across all models. This is mainly because the hyperparameter $N_Q$ is higher than the $\bar{N}_{n}$ on NLUE ($20 > 17.1$), affecting the training and evaluation on $D_{n}$.
\subsection{Ablation Study}
\paragraph{Multi-perspective Matching}
To evaluate the effectiveness of our Semantic Matching Module, we conduct further studies on individual components of our head matching. Table \ref{tab:component} shows that using only a single matching function is not sufficient to capture matching information between query and support samples. By aggregating all four matching methods, we observe a consistent improvement in both FSL and GFSL evaluations.

\begin{table}[t]
\centering
\caption{H-acc comparison on individual components of Semantic Matching module on SNIPS dataset.}
\resizebox{\columnwidth}{!}{%
\begin{tabular}{|c||c|c||c|c|}
\hline 
  & \multicolumn{2}{c||}{\textbf{1-shot}} & \multicolumn{2}{c|}{\textbf{5-shot}} \\
 \hline
 & \multicolumn{1}{c|}{\textbf{noneps}} & \multicolumn{1}{c||}{\textbf{eps}} & \multicolumn{1}{c|}{\textbf{noneps}} & \multicolumn{1}{c|}{\textbf{eps}} \\
\hline
Head-wise & 85.40 & 88.84  & 90.86 & 93.63 \\
\hline
Max-pooling & 85.54  & 88.87  & 90.79 & 93.63 \\
\hline
Attentive & 87.06 & 90.85  & 92.09  & 94.04 \\
\hline
Max-attentive & 87.37 & 90.87  & 92.18  & 94.37 \\
\hline
\textbf{Full Model}  & \textbf{88.29} & \textbf{91.64} & \textbf{92.21} &  \textbf{94.96} \\
\hline
\end{tabular}%
}
\label{tab:component}
\end{table}
\vspace*{-0.2cm}
\paragraph{Head Matching vs Word Matching} As introduced in Section 4, each head aims to extract a SC that covers a different aspect of a given sentence. To evaluate the effectiveness of head matching, we compare it with its corresponding word matching. In word matching, the hidden state embedding ($\mathbf{h}_i$) from Bi-LSTM is used for comparison rather than the head representation ($\mathbf{M}_i$). In addition, instead of head-wise matching, we compare each word forward and backward embedding of sentence $\mathbf{S}_k$ with the last (forward) and first (backward) embedding of sentence $\mathbf{Q}$ where $T_{q}$ denotes the last word in sentence $\mathbf{Q}$.
\vspace*{-0.35cm}
\begin{equation}
\begin{aligned}
& \overrightarrow{\mathbf{m}_i^{word\_wise}} = f_{m}(\overrightarrow{\mathbf{h}_{s_k}^i}, \overrightarrow{\mathbf{h}_{q}^{T_{q}}};\mathbf{W}^1) \\
& \overleftarrow{\mathbf{m}_i^{word\_wise}} = f_{m}(\overleftarrow{\mathbf{h}_{s_k}^i}, \overleftarrow{\mathbf{h}_{q}^1};\mathbf{W}^2)
\end{aligned}
\end{equation}

\begin{figure}[bt]
    \centering
    \includegraphics[trim={2.2cm 1.0cm 0.2cm 3.0cm},clip, width=\columnwidth]{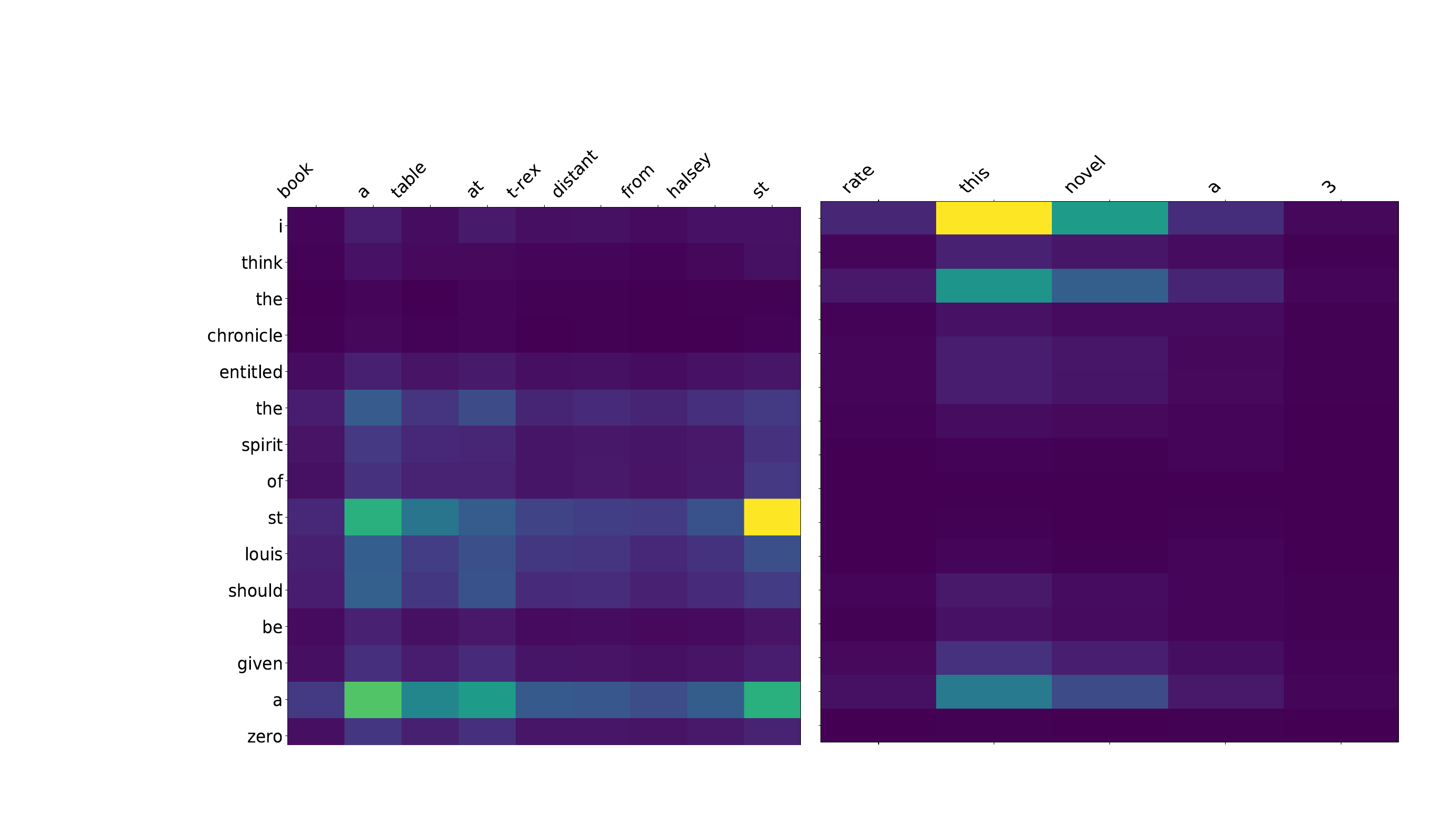}
    \caption{Word level matching. Y-axis denotes words of a sample query utterance \textit{``i think the chronicle entitled the spirit of st louis should be given a zero rating''} and X-axis (left) denote words of negative support utterance \textit{``book a table at t-rex distant from halsey st"} and X-axis (right) denotes positive support \textit{``rate this novel a 3"}. The label for query and positive support is ``Rate Book" and the negative support's label is ``Book Restaurant". The lighter color implies higher attention score.}
    \label{fig:word}
\end{figure}
\vspace*{-0.3cm}
Figure \ref{fig:word} illustrates an example when overly fine-grained matching sends the wrong matching signal, causing mis-classification for a query sample. Although \textit{``st''} exists in both query and negative support sample, it contains different meanings depending on contexts (\textit{``street''} vs \textit{``saint''}) and does not contribute to the correct intent \textit{``Rate Book"}. However, word matching assigns high matching score, leading to mis-classification of query sample as \textit{``Book Restaurant"} intent. As shown in the right part of Figure \ref{fig:word} word matching fails to identify indicative matching information with positive support sample (i.e. \textit{``rate"} vs \textit{``rating"}). 
This observation indicates that matching on the overly fine-grained word level semantics could lead to overfitting problems as only query samples of high word overlaps with support samples could yield high matching score. As utterances are diversely expressed, word-level semantic is insufficient to capture similarity between different utterances of the same intent.  

\begin{figure}[t]
    \centering
     \includegraphics[trim={10.0cm, 0.8cm, 5.0cm, 3.5cm},clip,width=\columnwidth]{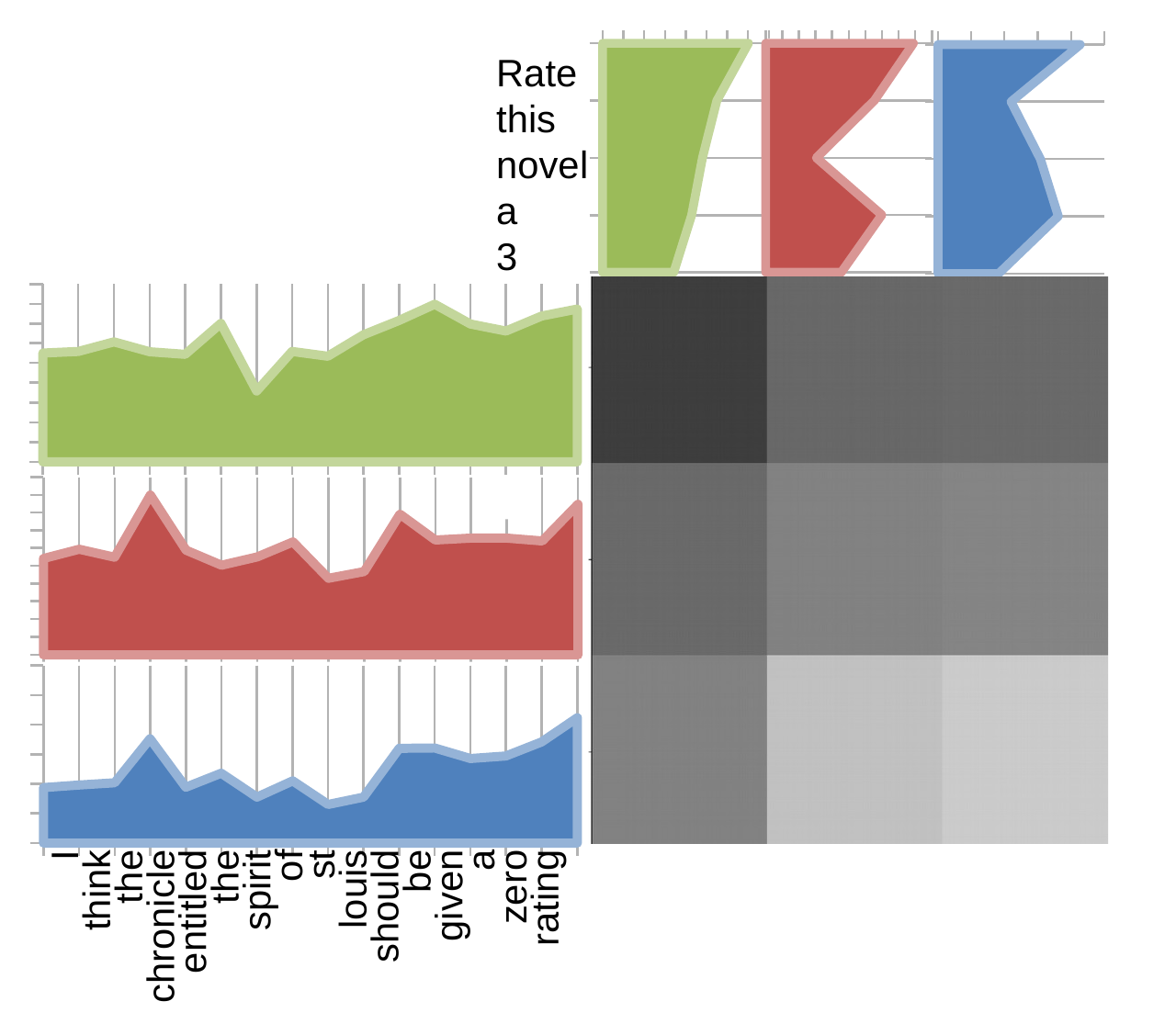}
    \caption{Head level matching between the same query and positive support utterance.  Y-axis denotes 3 heads extracted from query utterance labeled with the word distribution of each head. X-axis denotes 3 heads extracted from positive support utterance with similar label technique. Different curve colors are used to denote head indexes. The lighter color of each cell in 3x3 square matrix denotes the higher attention score.}
    \label{fig:head}
\end{figure}

On the other hand, when we use extracted heads for matching, as observed from Figure \ref{fig:head}, the importance of \textit{``st''} is significantly downplayed. Instead, query heads focus on extracting different aspects of the query: verb \textit{``should"}, \textit{``be"} (head 1), object target \textit{``chronicle"} (head 2), rating-related information \textit{ ``ratings"} (head 3). These key components are also captured in the positive support: target object(\textit{``novel"}) and rating keyword (\textit{``rate"}). As clearly indicated in Figure \ref{fig:head}, the head with color blue of query and positive support sample that both extract important rating-related keywords (\textit{``rating"} vs \textit{``rate"}) achieve high matching score.   

This observation confirms our intuitions (1) Each SC extracts essential high-level semantics of a given utterance, (2) Without sharing word-level similarity, essential keywords for intent label of query samples are extracted and matched with those from support samples (i.e. \textit{``rating"} vs \textit{``rate"}) via intermediate semantic component level. Further qualitative results in Table \ref{tab:headmatch} validate the effectiveness of head-vs-head matching as it outperforms its word matching counterpart in all evaluation scenarios. This is mainly because the semantic components extracted from SE effectively capture the most important words in the given utterances as observed in a sample query utterance, reducing the necessity to focus on matching irrelevant words.
\paragraph{Head Matching Regularization}
\begin{table}[ht]
\centering
\caption{H-accuracy evaluation on head matching vs word matching and regularization terms effectiveness on SNIPS dataset.}
\resizebox{\columnwidth}{!}{%
\begin{tabular}{|c||c|c||c|c||}
\hline 
  & \multicolumn{2}{c||}{\textbf{1-shot}} & \multicolumn{2}{c||}{\textbf{5-shot}} \\
 \hline
 & \multicolumn{1}{c|}{\textbf{noneps}} & \multicolumn{1}{c||}{\textbf{eps}} & \multicolumn{1}{c|}{\textbf{noneps}} & \multicolumn{1}{c||}{\textbf{eps}} \\
 \hline

Word Match & 85.94 & 90.87 & 90.11 & 93.22 \\
\hline
No $\mathcal{L}_{cross}$ & 87.58 & 91.2 & 92.06 & 94.84 \\
\hline
No $\mathcal{L}_{self}$ & 87.96 & 91.60 & 92.10  & 94.89 \\
\hline
No $\mathcal{L}_{uniform}$ & 87.65  & 91.17 & 92.09 & 94.92 \\
\hline
\textbf{Full Model} & \textbf{88.29} & \textbf{91.64} & \textbf{92.21} &  \textbf{94.96} \\
\hline
\end{tabular}%
}
\label{tab:headmatch}
\end{table}
\vspace*{-0.2cm}
As observed from Table \ref{tab:headmatch}, adding each additional regularization term boosts both GFSL and FSL performance. $\mathcal{L}_{cross}$ contributes most to the overall performance improvement. It is mainly due to its ability to align head distribution of samples with the same class label. Therefore, each extracted head could focus more on an indicative signal of the intent label.     
\section{Conclusions}
In this paper, we propose an effective Semantic Matching and Aggregation Network for few-shot intent detection. Semantic components extracted from multi-head self-attention capture higher level contextual information beyond the word level, enhancing model's generalizability towards both seen and novel intents, especially when utterances are diversely expressed. Comprehensive multi-perspective matching method thoroughly exploits the similarity between query and support samples for further robust representations. In this work, we also propose a more challenging but realistic non-episodic evaluation for both FSL and GFSL beyond traditional setting. Our model achieves the state-of-the-art performance in both evaluation settings for SNIPS and NLUE benchmark datasets. Further studies of more dynamic semantic extraction and effectively synthesized matching techniques are our desired future work.      
\section*{Acknowledgments}
We thank you reviewers for insightful feedback. This work is supported in part by NSF under grants III-1763325, III-1909323, and SaTC-1930941. 

We would like to acknowledge the use of the facilities of the High Performance Computing Division and High Performance Research and Development Group at the National Center for Atmospheric Research and the use of computational resources (doi:10.5065/D6RX99HX) at the NCAR-Wyoming Supercomputing Center provided by the National Science Foundation and the State of Wyoming, and supported by NCAR's Computational and Information Systems Laboratory.

\balance
\bibliography{emnlp2020}

\begin{thebibliography}{25}
\expandafter\ifx\csname natexlab\endcsname\relax\def\natexlab#1{#1}\fi

\bibitem[{Bojanowski et~al.(2017)Bojanowski, Grave, Joulin, and
  Mikolov}]{bojanowski2017enriching}
Piotr Bojanowski, Edouard Grave, Armand Joulin, and Tomas Mikolov. 2017.
\newblock Enriching word vectors with subword information.
\newblock \emph{Transactions of the Association for Computational Linguistics},
  5:135--146.

\bibitem[{Chen et~al.(2019)Chen, Liu, Kira, Wang, and
  Huang}]{chen2019closerfewshot}
Wei-Yu Chen, Yen-Cheng Liu, Zsolt Kira, Yu-Chiang Wang, and Jia-Bin Huang.
  2019.
\newblock A closer look at few-shot classification.
\newblock In \emph{International Conference on Learning Representations}.

\bibitem[{Devlin et~al.(2018)Devlin, Chang, Lee, and
  Toutanova}]{devlin2018bert}
Jacob Devlin, Ming-Wei Chang, Kenton Lee, and Kristina Toutanova. 2018.
\newblock Bert: Pre-training of deep bidirectional transformers for language
  understanding.
\newblock \emph{arXiv preprint arXiv:1810.04805}.

\bibitem[{Finn et~al.(2017)Finn, Abbeel, and Levine}]{finn2017model}
Chelsea Finn, Pieter Abbeel, and Sergey Levine. 2017.
\newblock Model-agnostic meta-learning for fast adaptation of deep networks.
\newblock In \emph{Proceedings of the 34th International Conference on Machine
  Learning-Volume 70}, pages 1126--1135. JMLR. org.

\bibitem[{Gao et~al.(2019)Gao, Han, Liu, and Sun}]{gao2019hybrid}
Tianyu Gao, Xu~Han, Zhiyuan Liu, and Maosong Sun. 2019.
\newblock Hybrid attention-based prototypical networks for noisy few-shot
  relation classification.
\newblock In \emph{Proceedings of the Thirty-Second AAAI Conference on
  Artificial Intelligence,(AAAI-19), New York, USA}.

\bibitem[{Gidaris and Komodakis(2018)}]{gidaris2018dynamic}
Spyros Gidaris and Nikos Komodakis. 2018.
\newblock Dynamic few-shot visual learning without forgetting.
\newblock In \emph{Proceedings of the IEEE Conference on Computer Vision and
  Pattern Recognition}, pages 4367--4375.

\bibitem[{Goo et~al.(2018)Goo, Gao, Hsu, Huo, Chen, Hsu, and
  Chen}]{goo2018slot}
Chih-Wen Goo, Guang Gao, Yun-Kai Hsu, Chih-Li Huo, Tsung-Chieh Chen, Keng-Wei
  Hsu, and Yun-Nung Chen. 2018.
\newblock Slot-gated modeling for joint slot filling and intent prediction.
\newblock In \emph{Proceedings of the 2018 Conference of the North American
  Chapter of the Association for Computational Linguistics: Human Language
  Technologies, Volume 2 (Short Papers)}, pages 753--757.

\bibitem[{Haihong et~al.(2019)Haihong, Niu, Chen, and Song}]{haihong2019novel}
E~Haihong, Peiqing Niu, Zhongfu Chen, and Meina Song. 2019.
\newblock A novel bi-directional interrelated model for joint intent detection
  and slot filling.
\newblock In \emph{Proceedings of the 57th Annual Meeting of the Association
  for Computational Linguistics}, pages 5467--5471.

\bibitem[{Lake et~al.(2011)Lake, Salakhutdinov, Gross, and
  Tenenbaum}]{lake2011one}
Brenden Lake, Ruslan Salakhutdinov, Jason Gross, and Joshua Tenenbaum. 2011.
\newblock One shot learning of simple visual concepts.
\newblock In \emph{Proceedings of the annual meeting of the cognitive science
  society}, volume~33.

\bibitem[{Lin et~al.(2017)Lin, Feng, Santos, Yu, Xiang, Zhou, and
  Bengio}]{lin2017structured}
Zhouhan Lin, Minwei Feng, Cicero Nogueira~dos Santos, Mo~Yu, Bing Xiang, Bowen
  Zhou, and Yoshua Bengio. 2017.
\newblock A structured self-attentive sentence embedding.
\newblock \emph{arXiv preprint arXiv:1703.03130}.

\bibitem[{Liu et~al.(2019{\natexlab{a}})Liu, Zhang, Fan, Fu, Li, Wu, and
  Lam}]{liu2019reconstructing}
Han Liu, Xiaotong Zhang, Lu~Fan, Xuandi Fu, Qimai Li, Xiao-Ming Wu, and
  Albert~YS Lam. 2019{\natexlab{a}}.
\newblock Reconstructing capsule networks for zero-shot intent classification.
\newblock In \emph{Proceedings of the 2019 Conference on Empirical Methods in
  Natural Language Processing and the 9th International Joint Conference on
  Natural Language Processing (EMNLP-IJCNLP)}, pages 4801--4811.

\bibitem[{Liu et~al.(2019{\natexlab{b}})Liu, Eshghi, Swietojanski, and
  Rieser}]{liu2019benchmarking}
Xingkun Liu, Arash Eshghi, Pawel Swietojanski, and Verena Rieser.
  2019{\natexlab{b}}.
\newblock Benchmarking natural language understanding services for building
  conversational agents.
\newblock In \emph{10th International Workshop on Spoken Dialogue Systems
  Technology 2019}.

\bibitem[{Mishra et~al.(2018)Mishra, Rohaninejad, Chen, and
  Abbeel}]{DBLP:conf/iclr/MishraR0A18}
Nikhil Mishra, Mostafa Rohaninejad, Xi~Chen, and Pieter Abbeel. 2018.
\newblock \href {https://openreview.net/forum?id=B1DmUzWAW} {A simple neural
  attentive meta-learner}.
\newblock In \emph{6th International Conference on Learning Representations,
  {ICLR} 2018, Vancouver, BC, Canada, April 30 - May 3, 2018, Conference Track
  Proceedings}. OpenReview.net.

\bibitem[{Pennington et~al.(2014)Pennington, Socher, and
  Manning}]{pennington2014glove}
Jeffrey Pennington, Richard Socher, and Christopher~D. Manning. 2014.
\newblock \href {http://www.aclweb.org/anthology/D14-1162} {Glove: Global
  vectors for word representation}.
\newblock In \emph{Empirical Methods in Natural Language Processing (EMNLP)},
  pages 1532--1543.

\bibitem[{Ravi and Larochelle(2017)}]{DBLP:conf/iclr/RaviL17}
Sachin Ravi and Hugo Larochelle. 2017.
\newblock \href {https://openreview.net/forum?id=rJY0-Kcll} {Optimization as a
  model for few-shot learning}.
\newblock In \emph{5th International Conference on Learning Representations,
  {ICLR} 2017, Toulon, France, April 24-26, 2017, Conference Track
  Proceedings}. OpenReview.net.

\bibitem[{Santoro et~al.(2016)Santoro, Bartunov, Botvinick, Wierstra, and
  Lillicrap}]{santoro2016meta}
Adam Santoro, Sergey Bartunov, Matthew Botvinick, Daan Wierstra, and Timothy
  Lillicrap. 2016.
\newblock Meta-learning with memory-augmented neural networks.
\newblock In \emph{International conference on machine learning}, pages
  1842--1850.

\bibitem[{Shi et~al.(2019)Shi, Salewski, Schiegg, Akata, and
  Welling}]{shi2019relational}
Xiahan Shi, Leonard Salewski, Martin Schiegg, Zeynep Akata, and Max Welling.
  2019.
\newblock Relational generalized few-shot learning.
\newblock \emph{arXiv preprint arXiv:1907.09557}.

\bibitem[{Snell et~al.(2017)Snell, Swersky, and Zemel}]{snell2017prototypical}
Jake Snell, Kevin Swersky, and Richard Zemel. 2017.
\newblock Prototypical networks for few-shot learning.
\newblock In \emph{Advances in Neural Information Processing Systems}, pages
  4077--4087.

\bibitem[{Sun et~al.(2019)Sun, Sun, Zhou, and Lv}]{sun-etal-2019-hierarchical}
Shengli Sun, Qingfeng Sun, Kevin Zhou, and Tengchao Lv. 2019.
\newblock \href {https://doi.org/10.18653/v1/D19-1045} {Hierarchical attention
  prototypical networks for few-shot text classification}.
\newblock In \emph{Proceedings of the 2019 Conference on Empirical Methods in
  Natural Language Processing and the 9th International Joint Conference on
  Natural Language Processing (EMNLP-IJCNLP)}, pages 476--485, Hong Kong,
  China. Association for Computational Linguistics.

\bibitem[{Sung et~al.(2018)Sung, Yang, Zhang, Xiang, Torr, and
  Hospedales}]{sung2018learning}
Flood Sung, Yongxin Yang, Li~Zhang, Tao Xiang, Philip~HS Torr, and Timothy~M
  Hospedales. 2018.
\newblock Learning to compare: Relation network for few-shot learning.
\newblock In \emph{Proceedings of the IEEE Conference on Computer Vision and
  Pattern Recognition}, pages 1199--1208.

\bibitem[{Vinyals et~al.(2016)Vinyals, Blundell, Lillicrap, Wierstra
  et~al.}]{vinyals2016matching}
Oriol Vinyals, Charles Blundell, Timothy Lillicrap, Daan Wierstra, et~al. 2016.
\newblock Matching networks for one shot learning.
\newblock In \emph{Advances in neural information processing systems}, pages
  3630--3638.

\bibitem[{Wang et~al.(2017)Wang, Hamza, and Florian}]{ijcai2017-579}
Zhiguo Wang, Wael Hamza, and Radu Florian. 2017.
\newblock \href {https://doi.org/10.24963/ijcai.2017/579} {Bilateral
  multi-perspective matching for natural language sentences}.
\newblock In \emph{Proceedings of the Twenty-Sixth International Joint
  Conference on Artificial Intelligence, {IJCAI-17}}, pages 4144--4150.

\bibitem[{Xia et~al.(2020)Xia, Zhang, Nguyen, Zhang, and Yu}]{xia2020cg}
Congying Xia, Chenwei Zhang, Hoang Nguyen, Jiawei Zhang, and Philip Yu. 2020.
\newblock Cg-bert: Conditional text generation with bert for generalized
  few-shot intent detection.
\newblock \emph{arXiv preprint arXiv:2004.01881}.

\bibitem[{Xia et~al.(2018)Xia, Zhang, Yan, Chang, and Philip}]{xia2018zero}
Congying Xia, Chenwei Zhang, Xiaohui Yan, Yi~Chang, and S~Yu Philip. 2018.
\newblock Zero-shot user intent detection via capsule neural networks.
\newblock In \emph{Proceedings of the 2018 Conference on Empirical Methods in
  Natural Language Processing}, pages 3090--3099.

\bibitem[{Ye and Ling(2019)}]{ye-ling-2019-multi}
Zhi-Xiu Ye and Zhen-Hua Ling. 2019.
\newblock \href {https://doi.org/10.18653/v1/P19-1277} {Multi-level matching
  and aggregation network for few-shot relation classification}.
\newblock In \emph{Proceedings of the 57th Annual Meeting of the Association
  for Computational Linguistics}, pages 2872--2881, Florence, Italy.
  Association for Computational Linguistics.

\end{thebibliography}
\bibliographystyle{acl_natbib}
\end{document}